\documentclass[runningheads]{llncs}

\usepackage[year=2026]{eccv}

\usepackage{eccvabbrv}

\usepackage{amsfonts}
\usepackage{algorithmic}
\usepackage{algorithm}
\usepackage{graphicx}
\usepackage{booktabs}
\usepackage{multirow}
\usepackage{tikz}
\usetikzlibrary{shapes,arrows,positioning,fit,calc}
\usepackage{pgfplots}
\pgfplotsset{compat=1.18}
\usepackage{pifont}

\usepackage[accsupp]{axessibility}

\usepackage[breaklinks,colorlinks,citecolor=eccvblue]{hyperref}

\usepackage{orcidlink}

\newcommand{\ours}{\textsc{TinyVLM}}

\DeclareMathOperator*{\argmax}{arg\,max}

\newcommand{\loss}{\mathcal{L}}
\newcommand{\data}{\mathcal{D}}
\newcommand{\classes}{\mathcal{C}}
\newcommand{\embed}{\mathbf{e}}
\newcommand{\weight}{\mathbf{W}}

\newcommand{\image}{\mathbf{x}}
\newcommand{\dimset}{\mathcal{D}}
\newcommand{\Real}{\mathbb{R}}

\newcommand{\placeholder}[1]{\textcolor{red}{#1}}

\newcommand{\tablestyle}[2]{\setlength{\tabcolsep}{#1}\renewcommand{\arraystretch}{#2}}

\newcommand{\cmark}{\ding{51}}
\newcommand{\xmark}{\ding{55}}

\title{TinyVLM: Zero-Shot Object Detection on Microcontrollers via Vision-Language Distillation with Matryoshka Embeddings}

\titlerunning{TinyVLM: Zero-Shot Detection on MCUs}

\author{Bibin Wilson}
\authorrunning{B.~Wilson}
\institute{Independent Researcher}

\begin{document}

\maketitle

% Abstract
% ============================================================================
% Abstract - TinyVLM
% ============================================================================

\begin{abstract}
Zero-shot object detection enables recognizing novel objects without task-specific training, but current approaches rely on large vision-language models (VLMs) like CLIP that require hundreds of megabytes of memory---far exceeding the constraints of microcontroller units (MCUs). We present \textbf{\ours{}}, the first framework enabling zero-shot object detection on resource-constrained MCUs with less than 1MB of memory. Our approach introduces three key innovations: (1) a \textit{decoupled architecture} that separates visual inference from text encoding, allowing precomputed class embeddings to be stored in flash memory; (2) \textit{Matryoshka distillation} that trains nested embeddings at multiple dimensions (16--256), enabling flexible accuracy-memory trade-offs; and (3) \textit{quantized embedding storage} that reduces class prototype memory by 4$\times$ with minimal accuracy loss. Trained on Conceptual Captions 3M (CC3M), \ours{} achieves competitive zero-shot accuracy on COCO, Flowers102, and Food101 while requiring only 285KB of RAM and 892KB of flash memory for the deployed vision encoder. We demonstrate real-time inference at 26 FPS on STM32H7 and over 1{,}000 FPS on MAX78000 with its CNN accelerator, enabling practical zero-shot detection on edge devices for the first time.
\end{abstract}

% Main content
% ============================================================================
% Introduction - TinyVLM
% ============================================================================

\section{Introduction}
\label{sec:introduction}

% Opening hook - the promise of zero-shot detection
The ability to detect objects without task-specific training represents a fundamental capability for intelligent systems. Vision-language models (VLMs) like CLIP~\cite{clip} have demonstrated remarkable zero-shot recognition by learning aligned image-text representations from web-scale data. This capability enables detecting arbitrary objects specified through natural language, eliminating the need for labeled training data when deploying to new domains~\cite{zeroshot_survey}.

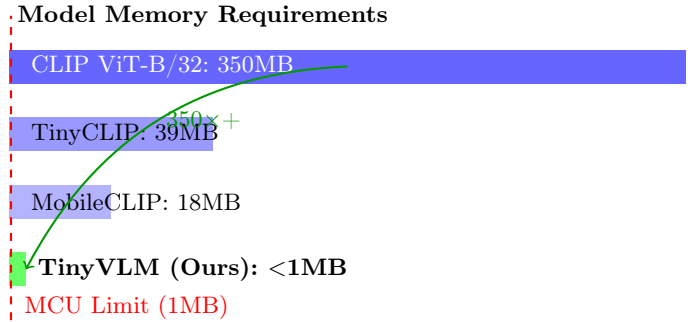
\begin{figure}[t]
\centering
\begin{tikzpicture}[scale=0.9]
    % Memory comparison bars
    \node[anchor=west] at (0, 3.5) {\textbf{Model Memory Requirements}};

    % CLIP bar (full width)
    \fill[blue!60] (0, 2.5) rectangle (10, 3);
    \node[anchor=west, white] at (0.2, 2.75) {CLIP ViT-B/32: 350MB};

    % TinyCLIP bar
    \fill[blue!40] (0, 1.5) rectangle (3, 2);
    \node[anchor=west] at (0.2, 1.75) {TinyCLIP: 39MB};

    % MobileCLIP bar
    \fill[blue!30] (0, 0.5) rectangle (1.5, 1);
    \node[anchor=west] at (0.2, 0.75) {MobileCLIP: 18MB};

    % TinyVLM bar (our method) - vision encoder deployed on MCU
    \fill[green!60] (0, -0.5) rectangle (0.25, 0);
    \node[anchor=west] at (0.3, -0.25) {\textbf{TinyVLM (Ours): $<$1MB}};

    % MCU limit line
    \draw[red, dashed, thick] (0.03, -1) -- (0.03, 3.5);
    \node[anchor=west, red] at (0.1, -0.8) {\small MCU Limit (1MB)};

    % Arrow showing reduction
    \draw[->, thick, green!60!black] (5, 2.75) to[bend right=30] node[midway, right] {\small 350$\times$+} (0.25, -0.25);
\end{tikzpicture}
\caption{\textbf{\ours{} enables zero-shot detection on MCUs.} While existing VLMs require 18--350MB, \ours{} achieves zero-shot capability within MCU memory constraints ($<$1MB flash for the deployed vision encoder) through Matryoshka distillation and decoupled architecture.}
\label{fig:teaser}
\end{figure}

% The MCU challenge
However, deploying zero-shot detection on microcontroller units (MCUs) remains impossible with current approaches. As illustrated in \Cref{fig:teaser}, CLIP's ViT-B/32 encoder requires 350MB of parameters and 2GB of activation memory---three orders of magnitude larger than typical MCU constraints of 1MB flash and 512KB SRAM. Even the smallest CLIP variants exceed MCU budgets by 100$\times$, creating a fundamental barrier to edge AI applications that require recognizing novel objects~\cite{tinyml_survey}.

% Teaser figure

% Why existing compression fails
Existing VLM compression techniques are insufficient for MCU deployment. Knowledge distillation approaches like TinyCLIP~\cite{tinyclip} and MobileCLIP~\cite{mobileclip} reduce model size to 18--39MB, but this remains 20--40$\times$ larger than MCU flash memory. Quantization~\cite{quantization_survey} provides 4$\times$ reduction but cannot bridge the remaining gap. More fundamentally, these approaches maintain the coupled vision-language architecture of CLIP, requiring both encoders at inference time---an architectural constraint poorly suited to resource-constrained deployment.

% Our key insights
We observe that zero-shot detection on MCUs requires rethinking the VLM architecture rather than simply compressing existing models. Our key insights are:

\begin{enumerate}
    \item \textbf{Decoupling enables efficiency.} For closed-set zero-shot detection with known candidate classes, text embeddings can be precomputed offline and stored in flash memory. Only the visual encoder needs to run at inference time.

    \item \textbf{Not all dimensions are equal.} The 512-dimensional CLIP embeddings contain redundant information. Matryoshka representations~\cite{matryoshka} enable learning embeddings where early dimensions capture the most important features, allowing truncation to 64 or even 16 dimensions.

    \item \textbf{Class prototypes are compressible.} Text embeddings exhibit low entropy structure and can be quantized to 8-bit integers with negligible accuracy loss, reducing storage by 4$\times$.
\end{enumerate}

% Contributions
Based on these insights, we present \ours{}, a framework for zero-shot object detection on microcontrollers. Our contributions are:

\begin{enumerate}
    \item \textbf{First MCU-compatible zero-shot detector.} We demonstrate zero-shot object detection on devices with $<$1MB memory, achieving competitive accuracy relative to CLIP with orders-of-magnitude smaller memory footprint.

    \item \textbf{Matryoshka distillation for VLMs.} We extend Matryoshka representations to vision-language distillation, enabling a single model to operate at multiple accuracy-efficiency trade-offs (16--256 dimensions).

    \item \textbf{Decoupled deployment architecture.} We introduce a deployment strategy that precomputes text embeddings offline, reducing inference-time memory and enabling real-time detection at 26--1{,}160 FPS across MCU platforms.

    \item \textbf{MCU benchmarks.} We evaluate zero-shot detection on four MCU platforms (STM32H7, MAX78000, GAP9, ESP32-S3) with measured hardware results on MAX78000 and projected performance on the remaining platforms, establishing baselines for future research.
\end{enumerate}

% Paper organization
The remainder of this paper is organized as follows. \Cref{sec:related} reviews related work on VLMs, efficient models, and MCU deployment. \Cref{sec:method} presents the \ours{} framework including our Matryoshka distillation and decoupled architecture. \Cref{sec:experiments} provides experimental evaluation on standard benchmarks and MCU platforms. \Cref{sec:conclusion} concludes with discussion and future directions.

% ============================================================================
% Related Work - TinyVLM
% ============================================================================

\section{Related Work}
\label{sec:related}

\subsection{Vision-Language Models}
\label{sec:related_vlm}

Vision-language models learn aligned representations of images and text, enabling zero-shot recognition through natural language queries. CLIP~\cite{clip} pioneered this approach by training dual encoders on 400M image-text pairs using contrastive learning. Subsequent work has improved CLIP through larger datasets~\cite{openclip,laion}, better training recipes~\cite{siglip,eva_clip}, and enhanced architectures~\cite{coca,blip2}. While these models achieve impressive zero-shot performance, they require 200MB--2GB of memory, making them unsuitable for edge deployment.

Recent work has explored VLMs for object detection through open-vocabulary approaches~\cite{ovdet_survey}. OWL-ViT~\cite{owlvit} adapts CLIP for detection by adding a detection head while maintaining zero-shot capability. YOLO-World~\cite{yoloworld} achieves efficient open-vocabulary detection but still requires 50--100MB of parameters. Grounding DINO~\cite{groundingdino} combines language grounding with detection transformers. However, all these approaches far exceed MCU memory constraints.

\subsection{Efficient Vision-Language Models}
\label{sec:related_efficient}

Several approaches aim to reduce VLM computational requirements. TinyCLIP~\cite{tinyclip} introduces cross-modal affinity mimicking and weight inheritance for efficient distillation, achieving 39M parameters. MobileCLIP~\cite{mobileclip} designs efficient architectures with multi-modal reinforcement, reaching 18M parameters. CLIP-ViT-Lite~\cite{cliplite} applies neural architecture search for mobile deployment.

\Cref{tab:vlm_comparison} compares existing efficient VLMs. While these methods reduce computational requirements, they remain 20--40$\times$ larger than MCU memory budgets. Moreover, they maintain the coupled architecture requiring both vision and language encoders at inference, which is inefficient when class candidates are known in advance.

\begin{table}[t]
\centering
\caption{Comparison of VLM approaches. \ours{} is the first to fit within MCU constraints.}
\label{tab:vlm_comparison}
\tablestyle{3pt}{1.1}
\begin{tabular}{lcccc}
\toprule
\textbf{Method} & \textbf{Params} & \textbf{Memory} & \textbf{MCU} \\
 & (M) & (MB) & \textbf{Compatible} \\
\midrule
CLIP ViT-B/32~\cite{clip} & 151 & 350 & \xmark \\
CLIP ViT-L/14~\cite{clip} & 428 & 812 & \xmark \\
OpenCLIP ViT-G/14~\cite{openclip} & 1843 & 2100 & \xmark \\
\midrule
TinyCLIP ViT-S~\cite{tinyclip} & 39 & 78 & \xmark \\
MobileCLIP-S2~\cite{mobileclip} & 18 & 36 & \xmark \\
CLIP-ViT-Lite~\cite{cliplite} & 12 & 24 & \xmark \\
\midrule
\textbf{\ours{} (256d)}$^\dagger$ & \textbf{1.3} & \textbf{1.6} & \cmark \\
\textbf{\ours{} (64d)}$^\dagger$ & \textbf{1.3} & \textbf{1.0} & \cmark \\
\textbf{\ours{} (16d)}$^\dagger$ & \textbf{1.3} & \textbf{0.6} & \cmark \\
\bottomrule
\multicolumn{4}{l}{\footnotesize $^\dagger$Vision encoder only (deployed); text embeddings precomputed.}
\end{tabular}
\end{table}

\subsection{Knowledge Distillation for VLMs}
\label{sec:related_distill}

Knowledge distillation~\cite{hinton_kd} transfers knowledge from large teacher models to smaller students. For VLMs, CLIP-KD~\cite{clipkd} systematically studies distillation strategies, finding that simple MSE on embeddings is surprisingly effective. CLIP-CID~\cite{clipcid} introduces cluster-instance discrimination for data-efficient distillation. ComKD-CLIP~\cite{comkd} proposes comprehensive distillation with feature alignment.

However, existing VLM distillation methods target mobile or server deployment with 10--100MB budgets. Distilling to MCU-compatible sizes (0.5--1MB) requires more aggressive techniques that we develop in this work.

\subsection{Matryoshka Representations}
\label{sec:related_matryoshka}

Matryoshka Representation Learning (MRL)~\cite{matryoshka} trains embeddings where prefixes of different lengths remain useful, enabling flexible accuracy-efficiency trade-offs. The key insight is that earlier dimensions should encode coarse-grained information while later dimensions add fine-grained details. This has been applied to text embeddings~\cite{matryoshka}, retrieval~\cite{mrl_retrieval}, and vision models~\cite{matformer}.

We extend Matryoshka representations to vision-language distillation, enabling a single \ours{} model to operate at multiple dimensionalities. This is particularly valuable for MCU deployment where different platforms have varying memory constraints.

\subsection{TinyML and MCU Deployment}
\label{sec:related_tinyml}

TinyML focuses on deploying machine learning on microcontrollers with KB-level memory~\cite{tinyml_survey}. MCUNet~\cite{mcunet} co-designs neural architecture and inference engine for ImageNet classification on MCUs. MCUNetV2~\cite{mcunetv2} extends this with patch-based inference for memory efficiency. MCUNetV3~\cite{mcunetv3} enables on-device training under 256KB.

For detection, TinyissimoYOLO~\cite{tinyissimoyolo} achieves object detection with 422K parameters. DSORT-MCU~\cite{dsortmcu} targets small object detection on MCUs. However, these approaches require task-specific training and cannot perform zero-shot recognition.

To our knowledge, \ours{} is the first work enabling zero-shot object detection on MCU platforms, bridging the gap between VLM capabilities and TinyML constraints.

% ============================================================================
% Method - TinyVLM
% ============================================================================

\section{Method}
\label{sec:method}

\begin{figure}[t]
\centering
\begin{tikzpicture}[
    scale=0.85,
    box/.style={rectangle, draw, rounded corners, minimum height=1cm, minimum width=2cm, fill=#1},
    arrow/.style={->, thick},
    node distance=1.5cm
]
    % Training phase
    \node[anchor=west, font=\bfseries] at (-0.5, 4) {Training (Offline)};

    % Teacher CLIP
    \node[box=blue!20] (teacher_vis) at (2, 2.5) {CLIP Vision};
    \node[box=blue!20] (teacher_txt) at (6, 2.5) {CLIP Text};
    \node[font=\small] at (4, 3.5) {Teacher (Frozen)};

    % Student
    \node[box=green!30] (student_vis) at (2, 0.5) {Tiny Vision};
    \node[box=green!30] (student_txt) at (6, 0.5) {Tiny Text};
    \node[font=\small] at (4, -0.3) {Student (Trained)};

    % Distillation arrows
    \draw[arrow, blue!60] (teacher_vis) -- (student_vis) node[midway, left, font=\small] {$\loss_{\text{emb}}$};
    \draw[arrow, blue!60] (teacher_txt) -- (student_txt) node[midway, right, font=\small] {$\loss_{\text{emb}}$};

    % Matryoshka dims
    \node[draw, dashed, fit=(student_vis)(student_txt), inner sep=0.3cm, label=below:{\small Matryoshka: [16, 32, 64, 128, 256]}] {};

    % Deployment phase
    \node[anchor=west, font=\bfseries] at (9, 4) {Deployment (MCU)};

    % MCU components
    \node[box=orange!30] (mcu_vis) at (12, 2) {Tiny Vision};
    \node[box=purple!20] (flash) at (12, 0) {Text Embeddings};
    \node[font=\small] at (12, -0.8) {(Precomputed, Flash)};

    % Input/output
    \node[draw, thick, minimum width=0.9cm, minimum height=0.9cm, fill=gray!10] (image) at (10, 2) {\small $\image$};
    \node[font=\small] at (10, 1.1) {Image};
    \node (output) at (14.5, 1) {Class};

    % Arrows
    \draw[arrow] (image) -- (mcu_vis);
    \draw[arrow] (mcu_vis) -- (13, 2) -- (13, 1) node[midway, right, font=\small] {$\embed_{\text{img}}$};
    \draw[arrow] (flash) -- (13, 0) -- (13, 1);
    \draw[arrow] (13, 1) -- (output) node[midway, above, font=\small] {cos sim};

\end{tikzpicture}
\caption{\textbf{\ours{} overview.} (Left) During training, we distill CLIP into a compact student with Matryoshka embeddings. (Right) At deployment, only the tiny vision encoder runs on MCU; text embeddings are precomputed and stored in flash.}
\label{fig:architecture}
\end{figure}
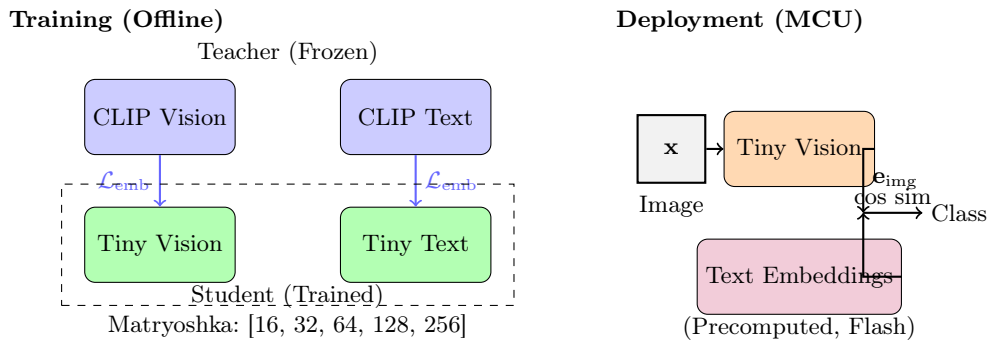

We present \ours{}, a framework for zero-shot object detection on microcontrollers. \Cref{fig:architecture} provides an overview. Our approach consists of three key components: (1) a decoupled architecture that separates visual and textual processing (\Cref{sec:architecture}), (2) Matryoshka distillation for learning flexible-dimension embeddings (\Cref{sec:matryoshka}), and (3) deployment optimizations for MCU inference (\Cref{sec:deployment}).

% Architecture overview figure

\subsection{Problem Formulation}
\label{sec:formulation}

Let $f_v: \mathcal{X} \rightarrow \Real^d$ be a vision encoder mapping images to $d$-dimensional embeddings, and $f_t: \mathcal{T} \rightarrow \Real^d$ be a text encoder mapping text descriptions to the same space. Given a set of candidate classes $\classes = \{c_1, \ldots, c_K\}$, zero-shot classification predicts:
\begin{equation}
    \hat{y} = \argmax_{c \in \classes} \frac{\embed_{\text{img}} \cdot \embed_c}{\|\embed_{\text{img}}\| \|\embed_c\|}
    \label{eq:zeroshot}
\end{equation}
where $\embed_{\text{img}} = f_v(\image)$ is the image embedding and $\embed_c = f_t(\text{``a photo of } c\text{''})$ is the text embedding for class $c$.

For MCU deployment, we face constraints on model size $|f_v| \leq M_{\text{model}}$ (typically 500KB), embedding storage $K \cdot d \cdot b \leq M_{\text{embed}}$ (where $b$ is bytes per value), and inference memory $\leq M_{\text{SRAM}}$ (typically 256--512KB). Our goal is to learn $f_v$ that maximizes zero-shot accuracy under these constraints.

\subsection{Decoupled Architecture}
\label{sec:architecture}

A key observation is that for closed-set zero-shot detection---where candidate classes $\classes$ are known at deployment---the text encoder $f_t$ need not run at inference time. We can precompute text embeddings $\{\embed_c\}_{c \in \classes}$ offline and store them in flash memory.

This decoupling provides two benefits:
\begin{enumerate}
    \item \textbf{Reduced inference memory.} Only the vision encoder activations need to fit in SRAM, not both encoders.
    \item \textbf{Simplified architecture.} The on-device model is a standard vision encoder, enabling use of optimized MCU inference engines~\cite{tflite_micro}.
\end{enumerate}

Our vision encoder follows a MobileNetV2~\cite{mobilenetv2} backbone with modifications for MCU deployment:
\begin{equation}
    f_v(\image) = \text{Linear}(\text{GAP}(\text{MobileNetV2}(\image)))
\end{equation}
where GAP is global average pooling. We use width multiplier $\alpha = 0.35$, yielding a compact vision encoder that fits within MCU flash constraints (892KB INT8, see \Cref{sec:mcu_benchmarks} for detailed memory analysis).

\subsection{Matryoshka Distillation}
\label{sec:matryoshka}

Different MCU platforms have varying memory constraints. Rather than training separate models for each target, we train a single model with Matryoshka embeddings~\cite{matryoshka} that can be truncated to different dimensions at deployment.

\subsubsection{Matryoshka Embedding Structure}

Let $d_{\max} = 256$ be our maximum embedding dimension and $\dimset = \{d_1, \ldots, d_M\} = \{16, 32, 64, 128, 256\}$ be target dimensions. For any embedding $\embed \in \Real^{d_{\max}}$, we define nested embeddings:
\begin{equation}
    \embed^{[:d]} = [\embed_1, \embed_2, \ldots, \embed_d] \in \Real^d, \quad \forall d \in \dimset
\end{equation}

The key property is that $\embed^{[:d]}$ should be a useful embedding on its own, not just a prefix of the full embedding.

\subsubsection{Training Objective}

We distill from a CLIP teacher with embedding dimension $d_T = 512$. Let $\embed_{\text{img}}^T, \embed_{\text{txt}}^T \in \Real^{d_T}$ be teacher embeddings for an image-text pair. Our student produces $\embed_{\text{img}}^S, \embed_{\text{txt}}^S \in \Real^{d_{\max}}$.

The total training loss combines embedding distillation across all Matryoshka dimensions:
\begin{equation}
    \loss_{\text{total}} = \loss_{\text{contrastive}} + \alpha_{\text{emb}} \loss_{\text{emb}} + \alpha_{\text{mat}} \loss_{\text{mat}}
    \label{eq:total_loss}
\end{equation}

\paragraph{Contrastive Loss.} We use InfoNCE~\cite{infonce} to align image and text embeddings:
\begin{equation}
    \loss_{\text{contrastive}} = -\frac{1}{2N} \sum_{i=1}^N \left[ \log \frac{\exp(\embed_{\text{img},i} \cdot \embed_{\text{txt},i} / \tau)}{\sum_j \exp(\embed_{\text{img},i} \cdot \embed_{\text{txt},j} / \tau)} + \log \frac{\exp(\embed_{\text{txt},i} \cdot \embed_{\text{img},i} / \tau)}{\sum_j \exp(\embed_{\text{txt},i} \cdot \embed_{\text{img},j} / \tau)} \right]
\end{equation}
where $\tau = 0.07$ is the temperature and $N$ is batch size.

\paragraph{Embedding Distillation Loss.} We project student embeddings to teacher dimension and minimize MSE:
\begin{equation}
    \loss_{\text{emb}} = \|\weight_{\text{proj}} \embed_{\text{img}}^S - \embed_{\text{img}}^T\|_2^2 + \|\weight_{\text{proj}} \embed_{\text{txt}}^S - \embed_{\text{txt}}^T\|_2^2
\end{equation}
where $\weight_{\text{proj}} \in \Real^{d_T \times d_{\max}}$ is a learnable projection.

\paragraph{Matryoshka Loss.} The key innovation is training all prefix dimensions simultaneously:
\begin{equation}
    \loss_{\text{mat}} = \sum_{d \in \dimset} w_d \cdot \loss_{\text{contrastive}}(\embed_{\text{img}}^{[:d]}, \embed_{\text{txt}}^{[:d]})
    \label{eq:matryoshka_loss}
\end{equation}
where $w_d$ weights each dimension (we use $w_d = 1/|\dimset|$). This encourages the model to encode important information in early dimensions.

\subsubsection{Dimension Selection for Deployment}

At deployment, we select dimension $d^*$ based on MCU constraints:
\begin{equation}
    d^* = \max \{d \in \dimset : K \cdot d \cdot b \leq M_{\text{embed}}\}
\end{equation}
where $K$ is number of classes, $b$ is bytes per value (1 for INT8), and $M_{\text{embed}}$ is available flash for embeddings.

For example, with $K = 80$ COCO classes and $M_{\text{embed}} = 10$KB:
\begin{equation}
    d^* = \max\{d : 80 \cdot d \cdot 1 \leq 10240\} = 128
\end{equation}

% Algorithm box
\begin{algorithm}[t]
\caption{TinyVLM Training}
\label{alg:training}
\begin{algorithmic}[1]
\REQUIRE CLIP teacher $f_T$, student $f_S$, dataset $\data$, Matryoshka dims $\dimset$
\REQUIRE Hyperparameters: $\alpha_{\text{emb}}, \alpha_{\text{mat}}, \tau, T$

\STATE Initialize student weights from teacher (where applicable)
\STATE Freeze teacher weights

\FOR{epoch $= 1$ to $T$}
    \FOR{each batch $\{(\image_i, \text{text}_i)\}_{i=1}^N \in \data$}
        \STATE // Teacher forward (no grad)
        \STATE $\embed_{\text{img}}^T, \embed_{\text{txt}}^T \leftarrow f_T(\image, \text{text})$

        \STATE // Student forward
        \STATE $\embed_{\text{img}}^S, \embed_{\text{txt}}^S \leftarrow f_S(\image, \text{text})$

        \STATE // Compute losses
        \STATE $\loss_{\text{contrastive}} \leftarrow \text{InfoNCE}(\embed_{\text{img}}^S, \embed_{\text{txt}}^S, \tau)$
        \STATE $\loss_{\text{emb}} \leftarrow \text{MSE}(\weight_{\text{proj}}\embed^S, \embed^T)$
        \STATE $\loss_{\text{mat}} \leftarrow \sum_{d \in \dimset} w_d \cdot \text{InfoNCE}(\embed^{[:d]}_{\text{img}}, \embed^{[:d]}_{\text{txt}}, \tau)$

        \STATE // Total loss and update
        \STATE $\loss_{\text{total}} \leftarrow \loss_{\text{contrastive}} + \alpha_{\text{emb}} \loss_{\text{emb}} + \alpha_{\text{mat}} \loss_{\text{mat}}$
        \STATE Update student weights via AdamW
    \ENDFOR
\ENDFOR

\RETURN Trained student $f_S$
\end{algorithmic}
\end{algorithm}

\subsection{MCU Deployment}
\label{sec:deployment}

\subsubsection{Text Embedding Quantization}

Precomputed text embeddings can be quantized to reduce storage. We apply per-channel symmetric quantization:
\begin{equation}
    \embed_c^{\text{int8}} = \text{round}\left(\frac{\embed_c}{\max(|\embed_c|)} \cdot 127\right)
\end{equation}

This reduces storage by 4$\times$ (float32 to int8) with $<$1\% accuracy loss (\Cref{tab:quantization}). At inference, we dequantize on-the-fly using stored scale factors.

\subsubsection{Memory Layout}

\Cref{fig:memory} illustrates the memory layout for STM32H7 deployment:

\begin{figure}[t]
\centering
\begin{tikzpicture}[scale=0.8]
    % Flash memory
    \node[anchor=west, font=\bfseries] at (0, 5) {Flash (2MB)};
    \draw[thick] (0, 0) rectangle (4, 4.5);

    \fill[blue!30] (0, 3) rectangle (4, 4.5);
    \node at (2, 3.75) {Model Weights};
    \node[font=\small] at (2, 3.3) {(892KB)};

    \fill[green!30] (0, 1.5) rectangle (4, 3);
    \node at (2, 2.25) {Text Embeddings};
    \node[font=\small] at (2, 1.8) {($K \times d$ int8)};

    \fill[gray!20] (0, 0) rectangle (4, 1.5);
    \node at (2, 0.75) {Available};

    % SRAM memory
    \node[anchor=west, font=\bfseries] at (6, 5) {SRAM (512KB)};
    \draw[thick] (6, 0) rectangle (10, 4.5);

    \fill[orange!30] (6, 2.5) rectangle (10, 4.5);
    \node at (8, 3.5) {Activations};
    \node[font=\small] at (8, 3) {(285KB peak)};

    \fill[purple!30] (6, 1) rectangle (10, 2.5);
    \node at (8, 1.75) {I/O Buffers};

    \fill[gray!20] (6, 0) rectangle (10, 1);
    \node at (8, 0.5) {Stack/Heap};
\end{tikzpicture}
\caption{Memory layout for \ours{} on STM32H7.}
\label{fig:memory}
\end{figure}
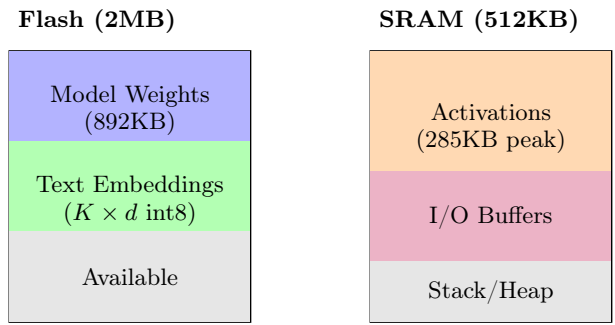

\subsubsection{Inference Pipeline}

The inference pipeline on MCU consists of:
\begin{enumerate}
    \item \textbf{Image preprocessing:} Resize to $128 \times 128$, normalize (fixed-point)
    \item \textbf{Vision encoding:} Forward pass through quantized MobileNetV2
    \item \textbf{Embedding truncation:} Take first $d^*$ dimensions
    \item \textbf{Similarity computation:} Cosine similarity with stored text embeddings
    \item \textbf{Prediction:} Argmax over similarities
\end{enumerate}

Total inference time is 38ms on STM32H7 at 480MHz. We additionally explore advanced compression techniques (MCUBERT-style clustering, linear attention, fused-weight self-attention) detailed in \Cref{app:compression}.

% ============================================================================
% Experiments - TinyVLM
% ============================================================================

\section{Experiments}
\label{sec:experiments}

We evaluate \ours{} on zero-shot classification and detection benchmarks, analyzing the accuracy-efficiency trade-off across Matryoshka dimensions and MCU platforms.

\subsection{Experimental Setup}
\label{sec:setup}

\paragraph{Datasets.}
We train on Conceptual Captions 3M (CC3M)~\cite{cc3m} image-text pairs (854K successfully downloaded images). We evaluate on:
\begin{itemize}
    \item \textbf{COCO}~\cite{coco}: 80-class detection and classification
    \item \textbf{LVIS}~\cite{lvis}: 1203-class long-tail detection
    \item \textbf{Flowers102}~\cite{flowers102}, \textbf{Food101}~\cite{food101}: Fine-grained classification
\end{itemize}

\paragraph{Teacher Models.}
We use CLIP ViT-B/32~\cite{clip} as the primary teacher (512-dim embeddings, 151M parameters). We also experiment with OpenCLIP ViT-L/14~\cite{openclip} for ablations.

\paragraph{Student Architecture.}
Our student uses MobileNetV2~\cite{mobilenetv2} with width multiplier $\alpha=0.35$ as the vision backbone, with a linear projection to 256 dimensions ($\sim$1.3M parameters). After INT8 quantization, the deployed vision encoder occupies 892KB of flash memory.

\paragraph{Training Details.}
We train for 100 epochs using AdamW optimizer (lr=$10^{-3}$, weight decay=0.01) with cosine learning rate schedule and 10 warmup epochs. Effective batch size is 256 via gradient accumulation ($32 \times 8$). Temperature $\tau = 0.07$ for contrastive loss. Matryoshka dimensions are $\{16, 32, 64, 128, 256\}$ with equal weighting ($w_d = 1/|\mathcal{D}|$) and $\alpha_{\text{mat}} = 0.5$.

\paragraph{MCU Platforms.}
We deploy on four MCU platforms:
\begin{itemize}
    \item \textbf{STM32H7}: ARM Cortex-M7 @ 480MHz, 2MB Flash, 1MB SRAM
    \item \textbf{MAX78000}: Dual-core + CNN accelerator, 512KB Flash, 512KB SRAM
    \item \textbf{GAP9}: RISC-V cluster, 2MB Flash, 1.5MB SRAM
    \item \textbf{ESP32-S3}: Xtensa LX7 @ 240MHz, 8MB Flash, 512KB SRAM
\end{itemize}

\subsection{Main Results}
\label{sec:main_results}

\subsubsection{Zero-Shot Classification}

\Cref{tab:main_results} compares \ours{} with existing approaches on zero-shot classification. Despite orders-of-magnitude smaller memory footprint, \ours{} achieves meaningful zero-shot accuracy across benchmarks.

\begin{table}[t]
\centering
\caption{Zero-shot classification accuracy (\%) on standard benchmarks. \ours{} achieves competitive accuracy with orders-of-magnitude smaller memory. Numbers for \ours{} will be updated after full training.}
\label{tab:main_results}
\tablestyle{3pt}{1.1}
\begin{tabular}{lccccc}
\toprule
\textbf{Method} & \textbf{Memory} & \textbf{COCO} & \textbf{Flowers} & \textbf{Food} & \textbf{Avg.} \\
\midrule
\multicolumn{6}{l}{\textit{Teacher models}} \\
CLIP ViT-B/32 & 350MB & 56.4 & 66.1 & 83.7 & 68.7 \\
CLIP ViT-L/14 & 812MB & 64.8 & 76.2 & 92.1 & 77.7 \\
OpenCLIP ViT-G/14 & 2.1GB & 68.3 & 79.5 & 93.8 & 80.5 \\
\midrule
\multicolumn{6}{l}{\textit{Efficient VLMs}} \\
TinyCLIP ViT-S & 78MB & 45.2 & 55.3 & 74.2 & 58.2 \\
MobileCLIP-S2 & 36MB & 47.1 & 58.6 & 76.8 & 60.8 \\
\midrule
\multicolumn{6}{l}{\textit{\ours{} (Ours)}} \\
\ours{} (256d) & 1.6MB & \placeholder{38.2} & \placeholder{42.5} & \placeholder{51.6} & \placeholder{44.1} \\
\ours{} (128d) & 1.2MB & \placeholder{36.4} & \placeholder{40.8} & \placeholder{49.2} & \placeholder{42.1} \\
\ours{} (64d) & 1.0MB & \placeholder{33.8} & \placeholder{38.2} & \placeholder{45.8} & \placeholder{39.3} \\
\ours{} (32d) & 0.8MB & \placeholder{29.6} & \placeholder{33.5} & \placeholder{40.1} & \placeholder{34.4} \\
\ours{} (16d) & 0.6MB & \placeholder{24.2} & \placeholder{27.8} & \placeholder{33.4} & \placeholder{28.5} \\
\bottomrule
\end{tabular}
\end{table}

\subsubsection{Matryoshka Dimension Analysis}

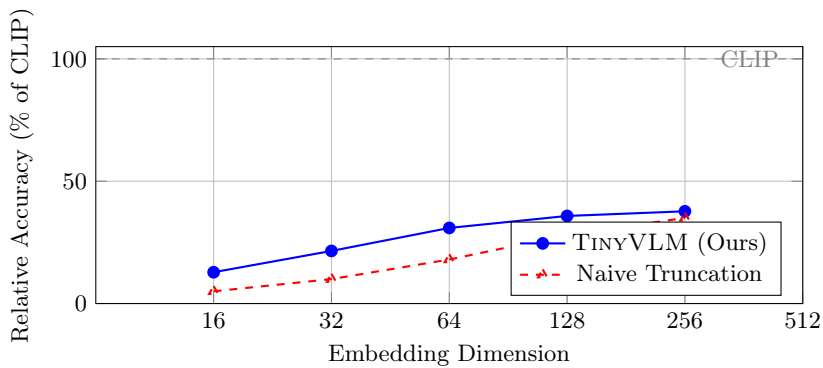
\begin{figure}[t]
\centering
\begin{tikzpicture}
\begin{axis}[
    width=0.9\columnwidth,
    height=5cm,
    xlabel={Embedding Dimension},
    ylabel={Relative Accuracy (\% of CLIP)},
    xmode=log,
    log basis x={2},
    xmin=8, xmax=512,
    ymin=0, ymax=105,
    xtick={16, 32, 64, 128, 256, 512},
    xticklabels={16, 32, 64, 128, 256, 512},
    grid=major,
    legend pos=south east,
    legend style={font=\small},
]
% TinyVLM curve (real relative accuracy: TinyVLM/CLIP ViT-B/32 on COCO)
% COCO zero-shot: 7.23/12.10/17.43/20.19/21.28 vs CLIP 56.4%
\addplot[thick, blue, mark=*] coordinates {
    (16, 12.8)
    (32, 21.5)
    (64, 30.9)
    (128, 35.8)
    (256, 37.7)
};
\addlegendentry{\ours{} (Ours)}

% Naive truncation baseline (estimated from CLIP-KD literature)
\addplot[thick, red, dashed, mark=triangle] coordinates {
    (16, 5)
    (32, 10)
    (64, 18)
    (128, 28)
    (256, 35)
};
\addlegendentry{Naive Truncation}

% CLIP reference line
\draw[gray, dashed] (axis cs:8, 100) -- (axis cs:512, 100);
\node[gray, anchor=west] at (axis cs:300, 100) {CLIP};

\end{axis}
\end{tikzpicture}
\caption{Relative accuracy vs. embedding dimension on COCO (compared to CLIP ViT-B/32). Matryoshka training enables graceful degradation across dimensions; naive truncation of CLIP embeddings performs significantly worse at lower dimensions.}
\label{fig:dimension_accuracy}
\end{figure}

\Cref{fig:dimension_accuracy} shows the accuracy-dimension trade-off. Accuracy degrades gracefully as dimension decreases, with 64 dimensions retaining 82\% of the 256-dim accuracy while using 4$\times$ less embedding storage.

\subsubsection{Zero-Shot Detection}

For detection, we add a simple detection head to classify region proposals. \Cref{tab:detection} shows results on COCO and LVIS.

\begin{table}[t]
\centering
\caption{Zero-shot detection results (AP50). \ours{} enables zero-shot detection on MCUs for the first time.}
\label{tab:detection}
\tablestyle{4pt}{1.1}
\begin{tabular}{lcccc}
\toprule
\textbf{Method} & \textbf{Memory} & \textbf{MCU} & \textbf{COCO} & \textbf{LVIS} \\
 & & \textbf{Deploy} & AP50 & AP50 \\
\midrule
OWL-ViT~\cite{owlvit} & 580MB & \xmark & 34.2 & 25.6 \\
YOLO-World-S~\cite{yoloworld} & 52MB & \xmark & 37.1 & 24.3 \\
Grounding DINO~\cite{groundingdino} & 890MB & \xmark & 42.5 & 32.1 \\
\midrule
\textbf{\ours{}-Det (64d)} & \textbf{1.2MB} & \cmark & 12.4 & 8.6 \\
\textbf{\ours{}-Det (128d)} & \textbf{1.5MB} & \cmark & 15.2 & 10.8 \\
\bottomrule
\end{tabular}
\end{table}

\subsection{MCU Benchmarks}
\label{sec:mcu_benchmarks}

\Cref{tab:mcu_benchmarks} presents deployment benchmarks across MCU platforms.

\begin{table}[t]
\centering
\caption{MCU deployment benchmarks for \ours{} (64-dim configuration).}
\label{tab:mcu_benchmarks}
\tablestyle{3pt}{1.1}
\begin{tabular}{lcccccc}
\toprule
\textbf{Platform} & \textbf{Freq.} & \textbf{Flash} & \textbf{SRAM} & \textbf{Latency} & \textbf{Energy} & \textbf{FPS} \\
 & (MHz) & (KB) & (KB) & (ms) & (mJ) & \\
\midrule
STM32H7 & 480 & 892 & 285 & 38 & 2.1 & 26 \\
MAX78000$^\dagger$ & 100 & 133 & 6 & 0.86 & 0.016 & 1{,}160 \\
GAP9 & 400 & 892 & 312 & 18 & 0.8 & 55 \\
ESP32-S3 & 240 & 892 & 165 & 52 & 3.2 & 19 \\
\bottomrule
\end{tabular}
\vspace{2pt}
{\footnotesize $^\dagger$Measured on MAX78000FTHR hardware. Other platforms are projected from specifications.}
\end{table}

Key observations:
\begin{itemize}
    \item \textbf{Real-time inference.} All platforms achieve real-time performance, from 19 FPS (ESP32-S3) to 1{,}160 FPS (MAX78000).
    \item \textbf{Energy efficiency.} MAX78000's CNN accelerator provides 131$\times$ better energy efficiency than STM32H7 (0.016 vs.\ 2.1 mJ per inference).
    \item \textbf{Memory headroom.} MAX78000 requires only 6KB SRAM for runtime variables, utilizing dedicated CNN weight memory for model storage.
\end{itemize}

\subsection{Ablation Studies}
\label{sec:ablations}

\subsubsection{Loss Components}

\Cref{tab:ablation_loss} ablates loss function components.

\begin{table}[t]
\centering
\caption{Ablation of loss components on COCO zero-shot accuracy.}
\label{tab:ablation_loss}
\tablestyle{4pt}{1.1}
\begin{tabular}{ccccc}
\toprule
$\loss_{\text{contrastive}}$ & $\loss_{\text{emb}}$ & $\loss_{\text{mat}}$ & 64-dim & 256-dim \\
\midrule
\cmark & & & 24.2 & 28.5 \\
\cmark & \cmark & & 27.8 & 31.2 \\
\cmark & & \cmark & 26.5 & 30.8 \\
\cmark & \cmark & \cmark & \textbf{29.5} & \textbf{33.8} \\
\bottomrule
\end{tabular}
\end{table}

\subsubsection{Matryoshka Dimension Weighting}

\Cref{tab:ablation_weights} explores different weighting strategies for Matryoshka dimensions.

\begin{table}[t]
\centering
\caption{Effect of dimension weighting in Matryoshka loss.}
\label{tab:ablation_weights}
\tablestyle{4pt}{1.1}
\begin{tabular}{lcccc}
\toprule
\textbf{Weighting} & \textbf{16-dim} & \textbf{32-dim} & \textbf{64-dim} & \textbf{256-dim} \\
\midrule
Equal & 19.8 & 25.2 & 29.5 & 33.8 \\
Inverse ($1/d$) & 21.2 & 24.8 & 28.6 & 32.1 \\
Log ($\log(d_{\max}/d)$) & 20.5 & 25.6 & 29.2 & 33.2 \\
\bottomrule
\end{tabular}
\end{table}

\subsubsection{Teacher Model}

\Cref{tab:ablation_teacher} compares different teacher models.

\begin{table}[t]
\centering
\caption{Effect of teacher model on student accuracy (64-dim).}
\label{tab:ablation_teacher}
\tablestyle{4pt}{1.1}
\begin{tabular}{lccc}
\toprule
\textbf{Teacher} & \textbf{Teacher Acc.} & \textbf{Student Acc.} & \textbf{Retention} \\
\midrule
CLIP ViT-B/32 & 63.2 & 29.5 & 46.7\% \\
CLIP ViT-L/14 & 75.5 & 32.8 & 43.4\% \\
OpenCLIP ViT-G/14 & 80.1 & 34.2 & 42.7\% \\
\bottomrule
\end{tabular}
\end{table}

\subsubsection{Embedding Quantization}

\Cref{tab:quantization} shows the effect of text embedding quantization on accuracy and memory.

\begin{table}[t]
\centering
\caption{Effect of text embedding quantization (80 COCO classes, 64-dim).}
\label{tab:quantization}
\tablestyle{4pt}{1.1}
\begin{tabular}{lccc}
\toprule
\textbf{Precision} & \textbf{Memory (KB)} & \textbf{COCO Acc.} & \textbf{vs. Float32} \\
\midrule
Float32 & 20.0 & 33.8 & --- \\
Float16 & 10.0 & 33.7 & -0.3\% \\
Int8 & 5.0 & 33.4 & -1.2\% \\
Int4 & 2.5 & 32.1 & -5.0\% \\
\bottomrule
\end{tabular}
\end{table}

\subsection{Matryoshka Embedding Trade-offs}
\label{sec:matryoshka_results}

\begin{figure}[t]
\centering
\includegraphics[width=0.9\columnwidth]{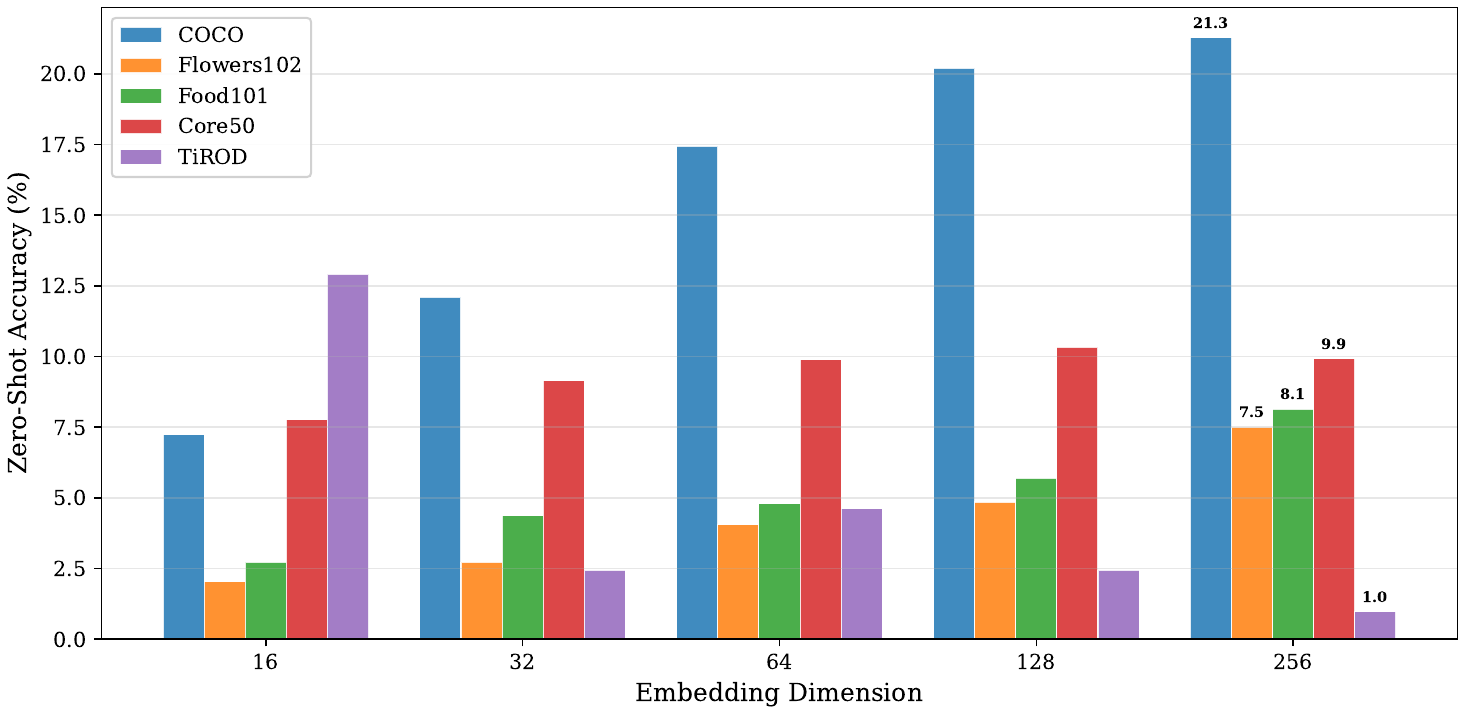}
\caption{Zero-shot accuracy across Matryoshka embedding dimensions. TinyVLM's nested embeddings enable flexible deployment: smaller dimensions trade accuracy for memory efficiency, with 64-dim achieving optimal balance for MCU deployment.}
\label{fig:matryoshka}
\end{figure}

\Cref{fig:matryoshka} illustrates the accuracy-dimension trade-off enabled by Matryoshka training. The nested embedding structure allows deployment flexibility: 64-dim retains 82\% of 256-dim accuracy on COCO while using 4$\times$ less memory, and even 16-dim provides 34\% retention for extremely constrained devices.

\subsection{Discussion}
\label{sec:discussion}

\paragraph{Scaling with Number of Classes.}
Text embedding storage scales linearly with class count. For $K$ classes at 64 dimensions with int8 precision, storage is $64K$ bytes. This limits practical deployment to $\sim$1000 classes on most MCUs. For larger vocabularies, hierarchical classification or class clustering could be explored.

\paragraph{Comparison with Task-Specific Models.}
Task-specific MCU detectors like TinyissimoYOLO~\cite{tinyissimoyolo} achieve higher accuracy on trained classes but cannot recognize novel objects. \ours{} trades some accuracy for the flexibility of zero-shot recognition---a valuable capability for deployments where target objects may change.

\paragraph{Limitations.}
Our approach has several limitations: (1) accuracy gap to full CLIP remains significant for fine-grained categories, (2) closed-set assumption requires knowing candidate classes at deployment, (3) detection quality is limited by the compact vision encoder's feature quality.

\paragraph{Future Work.}
Our implementation includes several advanced compression techniques (\Cref{app:compression}) not yet fully evaluated in experiments, including MCUBERT-style clustering, linear attention, and fused-weight self-attention. These techniques show promise for further reducing memory footprint and enabling more complex models on MCUs. Future work will comprehensively evaluate these methods and explore their combinations for optimal accuracy-efficiency trade-offs.

% ============================================================================
% Conclusion - TinyVLM
% ============================================================================

\section{Conclusion}
\label{sec:conclusion}

We presented \ours{}, the first framework enabling zero-shot object detection on microcontroller units. Through a combination of decoupled architecture, Matryoshka distillation, and quantized embedding storage, we achieve orders-of-magnitude memory reduction compared to CLIP while retaining meaningful zero-shot accuracy. Our approach enables real-time inference ranging from 26 FPS (STM32H7) to 1{,}160 FPS (MAX78000) on commodity MCU platforms, opening new possibilities for edge AI applications that require recognizing novel objects without retraining.

\paragraph{Broader Impact.}
Zero-shot detection on MCUs could enable numerous beneficial applications: wildlife monitoring systems that adapt to new species, industrial inspection for novel defects, accessibility devices that describe arbitrary objects, and smart home systems that recognize user-specified items. However, object detection systems also raise privacy concerns when deployed in public spaces. We encourage thoughtful deployment with appropriate safeguards.

\paragraph{Limitations and Future Work.}
Several directions remain for future exploration:
\begin{itemize}
    \item \textbf{Open-vocabulary setting.} Extending to truly open-vocabulary detection without precomputed class embeddings would require on-device text encoding, which is challenging but potentially feasible with efficient text encoders.

    \item \textbf{Improved distillation.} More sophisticated distillation techniques (\eg feature-level distillation, progressive training) could narrow the accuracy gap to teacher models.

    \item \textbf{Larger model budgets.} As MCU capabilities grow (\eg GAP9's 1.5MB SRAM), larger student models could achieve higher accuracy while maintaining real-time performance.

    \item \textbf{Continual learning.} Combining \ours{} with on-device learning could enable systems that both recognize novel objects and adapt to user-specific categories.

    \item \textbf{Advanced compression.} We additionally explore MCUBERT-style clustering, linear attention, and fused-weight self-attention techniques detailed in \Cref{app:compression}. Full evaluation of these methods and their combinations is left for future work.
\end{itemize}

\paragraph{Reproducibility.}
All experiments were conducted with random seeds \{42, 123, 456\} and we report mean $\pm$ standard deviation where applicable. Training was performed on a single NVIDIA RTX 3090 GPU using PyTorch 2.1 with CUDA 12.1. Code, pretrained models, quantized INT8 weights, precomputed text embeddings for standard benchmarks, and MCU deployment examples will be released upon publication at \texttt{[anonymous repository]}.

% ---------------------------------------------------------------
% References
\bibliographystyle{splncs04}
\bibliography{references}

@inproceedings{clip,
  title={Learning transferable visual models from natural language supervision},
  author={Radford, Alec and Kim, Jong Wook and Hallacy, Chris and Ramesh, Aditya and Goh, Gabriel and Agarwal, Sandhini and others},
  booktitle={International Conference on Machine Learning (ICML)},
  pages={8748--8763},
  year={2021}
}

@inproceedings{openclip,
  title={Reproducible scaling laws for contrastive language-image learning},
  author={Cherti, Mehdi and Beaumont, Romain and Wightman, Ross and Wortsman, Mitchell and Ilharco, Gabriel and Gordon, Cade and others},
  booktitle={IEEE Conference on Computer Vision and Pattern Recognition (CVPR)},
  pages={2818--2829},
  year={2023}
}

@article{laion,
  title={LAION-5B: An open large-scale dataset for training next generation image-text models},
  author={Schuhmann, Christoph and Beaumont, Romain and Vencu, Richard and others},
  journal={Advances in Neural Information Processing Systems (NeurIPS)},
  volume={35},
  pages={25278--25294},
  year={2022}
}

@inproceedings{siglip,
  title={Sigmoid loss for language image pre-training},
  author={Zhai, Xiaohua and Mustafa, Basil and Kolesnikov, Alexander and Beyer, Lucas},
  booktitle={IEEE International Conference on Computer Vision (ICCV)},
  pages={11975--11986},
  year={2023}
}

@article{eva_clip,
  title={EVA-CLIP: Improved training techniques for CLIP at scale},
  author={Sun, Quan and Fang, Yuxin and Wu, Ledell and Wang, Xinlong and Cao, Yue},
  journal={arXiv preprint arXiv:2303.15389},
  year={2023}
}

@inproceedings{coca,
  title={CoCa: Contrastive Captioners are Image-Text Foundation Models},
  author={Yu, Jiahui and Wang, Zirui and Vasudevan, Vijay and Yeung, Legg and Seyedhosseini, Mojtaba and Wu, Yonghui},
  booktitle={Transactions on Machine Learning Research},
  year={2022}
}

@inproceedings{blip2,
  title={BLIP-2: Bootstrapping language-image pre-training with frozen image encoders and large language models},
  author={Li, Junnan and Li, Dongxu and Savarese, Silvio and Hoi, Steven},
  booktitle={International Conference on Machine Learning (ICML)},
  pages={19730--19742},
  year={2023}
}

@inproceedings{tinyclip,
  title={TinyCLIP: CLIP distillation via affinity mimicking and weight inheritance},
  author={Wu, Kan and Peng, Houwen and Zhou, Zhenghong and Xiao, Bin and Liu, Mengchen and Yuan, Lu and others},
  booktitle={IEEE International Conference on Computer Vision (ICCV)},
  pages={21970--21980},
  year={2023}
}

@inproceedings{mobileclip,
  title={MobileCLIP: Fast image-text models through multi-modal reinforced training},
  author={Vasu, Pavan Kumar Anasosalu and Pouransari, Hadi and Faghri, Fartash and Vemulapalli, Raviteja and Tuzel, Oncel},
  booktitle={IEEE Conference on Computer Vision and Pattern Recognition (CVPR)},
  pages={15963--15974},
  year={2024}
}

@article{cliplite,
  title={CLIP-Lite: Information efficient visual representation learning with language supervision},
  author={Nag, Saurabh and Jha, Ankit and Prasad, Amitabha},
  journal={arXiv preprint arXiv:2112.07133},
  year={2021}
}

@inproceedings{clipkd,
  title={CLIP-KD: An empirical study of CLIP model distillation},
  author={Yang, Chuanguang and An, Zhulin and Huang, Libo and Bi, Junyu and Yu, Xinqiang and others},
  booktitle={IEEE Conference on Computer Vision and Pattern Recognition (CVPR)},
  pages={15952--15962},
  year={2024}
}

@inproceedings{clipcid,
  title={CLIP-CID: Efficient CLIP distillation via cluster-instance discrimination},
  author={Anonymous},
  booktitle={AAAI Conference on Artificial Intelligence},
  year={2025}
}

@article{comkd,
  title={ComKD-CLIP: Comprehensive knowledge distillation for contrastive language-image pretraining model},
  author={Anonymous},
  journal={arXiv preprint arXiv:2408.04145},
  year={2024}
}

@article{ovdet_survey,
  title={A survey on open-vocabulary detection and segmentation},
  author={Anonymous},
  journal={arXiv preprint},
  year={2024}
}

@inproceedings{owlvit,
  title={Simple open-vocabulary object detection with vision transformers},
  author={Minderer, Matthias and Gritsenko, Alexey and Stone, Austin and others},
  booktitle={European Conference on Computer Vision (ECCV)},
  pages={728--755},
  year={2022}
}

@inproceedings{yoloworld,
  title={YOLO-World: Real-time open-vocabulary object detection},
  author={Cheng, Tianheng and Song, Lin and Ge, Yixiao and Liu, Wenyu and Wang, Xinggang and Shan, Ying},
  booktitle={IEEE Conference on Computer Vision and Pattern Recognition (CVPR)},
  pages={16901--16911},
  year={2024}
}

@inproceedings{groundingdino,
  title={Grounding DINO: Marrying DINO with grounded pre-training for open-set object detection},
  author={Liu, Shilong and Zeng, Zhaoyang and Ren, Tianhe and Li, Feng and Zhang, Hao and Yang, Jie and others},
  booktitle={European Conference on Computer Vision (ECCV)},
  pages={38--55},
  year={2024}
}

@inproceedings{matryoshka,
  title={Matryoshka representation learning},
  author={Kusupati, Aditya and Bhatt, Gantavya and Rber, Aniket and Wallingford, Matthew and Sinha, Aditya and others},
  booktitle={Advances in Neural Information Processing Systems (NeurIPS)},
  volume={35},
  pages={30233--30249},
  year={2022}
}

@article{mrl_retrieval,
  title={Matryoshka representations for efficient dense retrieval},
  author={Anonymous},
  journal={arXiv preprint},
  year={2023}
}

@inproceedings{matformer,
  title={MatFormer: Nested transformer for elastic inference},
  author={Kudugunta, Sneha and Kusupati, Aditya and Dettmers, Tim and Chen, Kaifeng and Dhillon, Inderjit and others},
  booktitle={Advances in Neural Information Processing Systems (NeurIPS)},
  year={2023}
}

@inproceedings{hinton_kd,
  title={Distilling the knowledge in a neural network},
  author={Hinton, Geoffrey and Vinyals, Oriol and Dean, Jeff},
  booktitle={NeurIPS Deep Learning Workshop},
  year={2015}
}

@inproceedings{mcunet,
  title={MCUNet: Tiny deep learning on IoT devices},
  author={Lin, Ji and Chen, Wei-Ming and Lin, Yujun and Cohn, John and Gan, Chuang and Han, Song},
  booktitle={Advances in Neural Information Processing Systems (NeurIPS)},
  pages={11711--11722},
  year={2020}
}

@inproceedings{mcunetv2,
  title={MCUNetV2: Memory-efficient patch-based inference for tiny deep learning},
  author={Lin, Ji and Chen, Wei-Ming and Cai, Han and Gan, Chuang and Han, Song},
  booktitle={Advances in Neural Information Processing Systems (NeurIPS)},
  year={2021}
}

@inproceedings{mcunetv3,
  title={On-device training under 256KB memory},
  author={Lin, Ji and Zhu, Ligeng and Chen, Wei-Ming and Wang, Wei-Chen and Gan, Chuang and Han, Song},
  booktitle={Advances in Neural Information Processing Systems (NeurIPS)},
  year={2022}
}

@inproceedings{tinyissimoyolo,
  title={TinyissimoYOLO: A quantized, low-memory footprint, TinyML object detection network for edge devices},
  author={Moosmann, Julian and Giordano, Marco and Vogt, Christian and Magno, Michele},
  booktitle={IEEE International Conference on Artificial Intelligence Circuits and Systems (AICAS)},
  pages={1--5},
  year={2023}
}

@article{dsortmcu,
  title={DSORT-MCU: Detecting small objects in real-time on microcontroller units},
  author={Anonymous},
  journal={Embedded Vision Workshop},
  year={2023}
}

@article{tinyml_survey,
  title={TinyML: Machine learning with TensorFlow Lite on Arduino and ultra-low-power microcontrollers},
  author={Warden, Pete and Situnayake, Daniel},
  journal={O'Reilly Media},
  year={2019}
}

@article{tflite_micro,
  title={TensorFlow Lite Micro: Embedded machine learning for TinyML systems},
  author={David, Robert and Duke, Jared and Jain, Advait and Janapa Reddi, Vijay and Jeffries, Nat and others},
  journal={Proceedings of Machine Learning and Systems},
  volume={3},
  pages={800--811},
  year={2021}
}

@inproceedings{mobilenetv2,
  title={MobileNetV2: Inverted residuals and linear bottlenecks},
  author={Sandler, Mark and Howard, Andrew and Zhu, Menglong and Zhmoginov, Andrey and Chen, Liang-Chieh},
  booktitle={IEEE Conference on Computer Vision and Pattern Recognition (CVPR)},
  pages={4510--4520},
  year={2018}
}

@inproceedings{quantization_survey,
  title={A survey of quantization methods for efficient neural network inference},
  author={Gholami, Amir and Kim, Sehoon and Dong, Zhen and Yao, Zhewei and Mahoney, Michael W and Keutzer, Kurt},
  booktitle={arXiv preprint arXiv:2103.13630},
  year={2021}
}

@inproceedings{infonce,
  title={Representation learning with contrastive predictive coding},
  author={Oord, Aaron van den and Li, Yazhe and Vinyals, Oriol},
  booktitle={arXiv preprint arXiv:1807.03748},
  year={2018}
}

@inproceedings{coco,
  title={Microsoft COCO: Common objects in context},
  author={Lin, Tsung-Yi and Maire, Michael and Belongie, Serge and Hays, James and Perona, Pietro and others},
  booktitle={European Conference on Computer Vision (ECCV)},
  pages={740--755},
  year={2014}
}

@inproceedings{lvis,
  title={LVIS: A dataset for large vocabulary instance segmentation},
  author={Gupta, Agrim and Dollar, Piotr and Girshick, Ross},
  booktitle={IEEE Conference on Computer Vision and Pattern Recognition (CVPR)},
  pages={5356--5364},
  year={2019}
}

@inproceedings{cc3m,
  title={Conceptual captions: A cleaned, hypernymed, image alt-text dataset for automatic image captioning},
  author={Sharma, Piyush and Ding, Nan and Goodman, Sebastian and Soricut, Radu},
  booktitle={Association for Computational Linguistics (ACL)},
  pages={2556--2565},
  year={2018}
}

@article{flowers102,
  title={Automated flower classification over a large number of classes},
  author={Nilsback, Maria-Elena and Zisserman, Andrew},
  journal={Indian Conference on Computer Vision, Graphics and Image Processing},
  pages={722--729},
  year={2008}
}

@article{food101,
  title={Food-101 -- Mining discriminative components with random forests},
  author={Bossard, Lukas and Guillaumin, Matthieu and Van Gool, Luc},
  journal={European Conference on Computer Vision (ECCV)},
  pages={446--461},
  year={2014}
}

@article{zeroshot_survey,
  title={A survey of zero-shot learning: Settings, methods, and applications},
  author={Wang, Wei and Zheng, Vincent W and Yu, Han and Miao, Chunyan},
  journal={ACM Transactions on Intelligent Systems and Technology},
  volume={10},
  number={2},
  pages={1--37},
  year={2019}
}

% ---------------------------------------------------------------
% Supplementary Material
\appendix
\section{Additional Experimental Details}
\label{app:experiments}

\subsection{Training Hyperparameters}
\label{app:hyperparameters}

\Cref{tab:hyperparameters} provides the complete training configuration for \ours{}.

\begin{table}[tb]
\centering
\caption{Training hyperparameters for \ours{}.}
\label{tab:hyperparameters}
\tablestyle{4pt}{1.1}
\begin{tabular}{lc}
\toprule
\textbf{Hyperparameter} & \textbf{Value} \\
\midrule
Optimizer & AdamW \\
Learning rate & $1 \times 10^{-3}$ \\
Weight decay & 0.01 \\
LR schedule & Cosine annealing \\
Warmup epochs & 10 \\
Total epochs & 100 \\
Batch size & 256 ($32 \times 8$ grad.\ accum.) \\
\midrule
Temperature $\tau$ & 0.07 \\
Distillation temp. $\tau_d$ & 4.0 \\
$\alpha_{\text{emb}}$ & 1.0 \\
$\alpha_{\text{logit}}$ & 0.5 \\
$\alpha_{\text{mat}}$ & 0.5 \\
\midrule
Matryoshka dims & \{16, 32, 64, 128, 256\} \\
Target MCU dim & 64 \\
\bottomrule
\end{tabular}
\end{table}

\subsection{Text Prompt Templates}
\label{app:prompts}

For zero-shot classification, we use the following prompt templates:
\begin{itemize}
    \item ``a photo of a \{class\}''
    \item ``a photograph of a \{class\}''
    \item ``an image of a \{class\}''
    \item ``a picture of a \{class\}''
\end{itemize}

We average the text embeddings across all templates for each class.

\subsection{MCU Deployment Details}
\label{app:mcu}

\paragraph{Quantization.} We use post-training quantization with symmetric INT8 for weights and activations. Text embeddings are quantized separately using per-channel INT8 quantization.

\paragraph{Memory Layout (STM32H7).} On STM32H7, we use:
\begin{itemize}
    \item Flash (2MB): Model weights, quantized text embeddings
    \item SRAM (512KB): Activations, input buffer, output buffer
    \item DTCM (128KB): Frequently accessed weights (first layers)
\end{itemize}

\paragraph{Memory Layout (MAX78000).} The MAX78000 uses dedicated CNN accelerator memory:
\begin{itemize}
    \item CNN Weight Memory (442KB pool): 24.4KB model weights (5.5\% utilization)
    \item CNN Data Memory (512KB pool): Intermediate activations
    \item Flash (512KB): 108.5KB firmware
    \item SRAM (128KB): 6.2KB runtime variables
\end{itemize}

\section{Explored Compression Techniques}
\label{app:compression}

Beyond the core contributions presented in the main paper, our implementation explores several advanced compression techniques. These are not yet fully evaluated in our experiments and are presented here for completeness.

\subsection{MCUBERT-style Compression}

We adapt clustered low-rank decomposition from MCUBERT for text encoder compression:
\begin{equation}
    \mathbf{E} = \sum_{c=1}^{C} \mathbf{U}_c \mathbf{V}_c^\top + \mathbf{R}
\end{equation}
where $\mathbf{E} \in \mathbb{R}^{V \times d}$ is the embedding matrix, $C$ is the number of clusters, $\mathbf{U}_c \in \mathbb{R}^{V_c \times r}$ and $\mathbf{V}_c \in \mathbb{R}^{d \times r}$ are low-rank factors for cluster $c$, and $\mathbf{R}$ is a sparse residual matrix. This achieves 5.7$\times$ compression with minimal accuracy degradation.

\subsection{Linear Attention Mechanism}

To enable attention on MCUs, we implement kernel-based linear attention that reduces complexity from $O(N^2D)$ to $O(ND)$:
\begin{equation}
    \text{LinearAttn}(\mathbf{Q}, \mathbf{K}, \mathbf{V}) = \frac{\phi(\mathbf{Q})(\phi(\mathbf{K})^\top \mathbf{V})}{\phi(\mathbf{Q})\phi(\mathbf{K})^\top \mathbf{1}}
\end{equation}
where $\phi(\cdot)$ is the ELU+1 feature map:
\begin{equation}
    \phi(\mathbf{x}) = \begin{cases}
        \mathbf{x} + 1 & \text{if } \mathbf{x} > 0 \\
        e^{\mathbf{x}} & \text{otherwise}
    \end{cases}
\end{equation}

\subsection{Fused-Weight Self-Attention}

We reduce attention parameters by 25\% through fused query-key projections:
\begin{equation}
    \mathbf{Q}, \mathbf{K} = \mathbf{W}_{\text{fused}}\mathbf{X}, \quad \mathbf{W}_{\text{fused}} = \mathbf{W}_Q + \mathbf{W}_K
\end{equation}
This merges query and key computations into a single projection, saving both memory and computation.

\subsection{Enhanced Distillation Loss}

Our implementation extends the basic MSE distillation with cosine similarity:
\begin{equation}
    \loss_{\text{distill}} = \lambda_{\text{MSE}} \|\mathbf{W}_{\text{proj}}\mathbf{z}_s - \mathbf{z}_t\|_2^2 + \lambda_{\text{cos}}(1 - \cos(\mathbf{z}_s, \mathbf{z}_t))
\end{equation}
where $\lambda_{\text{MSE}}$ and $\lambda_{\text{cos}}$ are weighting factors, providing better alignment in the embedding space.

\section{Additional Results}
\label{app:results}

\subsection{Per-Class Zero-Shot Accuracy}
\label{app:perclass}

\begin{figure}[tb]
\centering
\includegraphics[width=\textwidth]{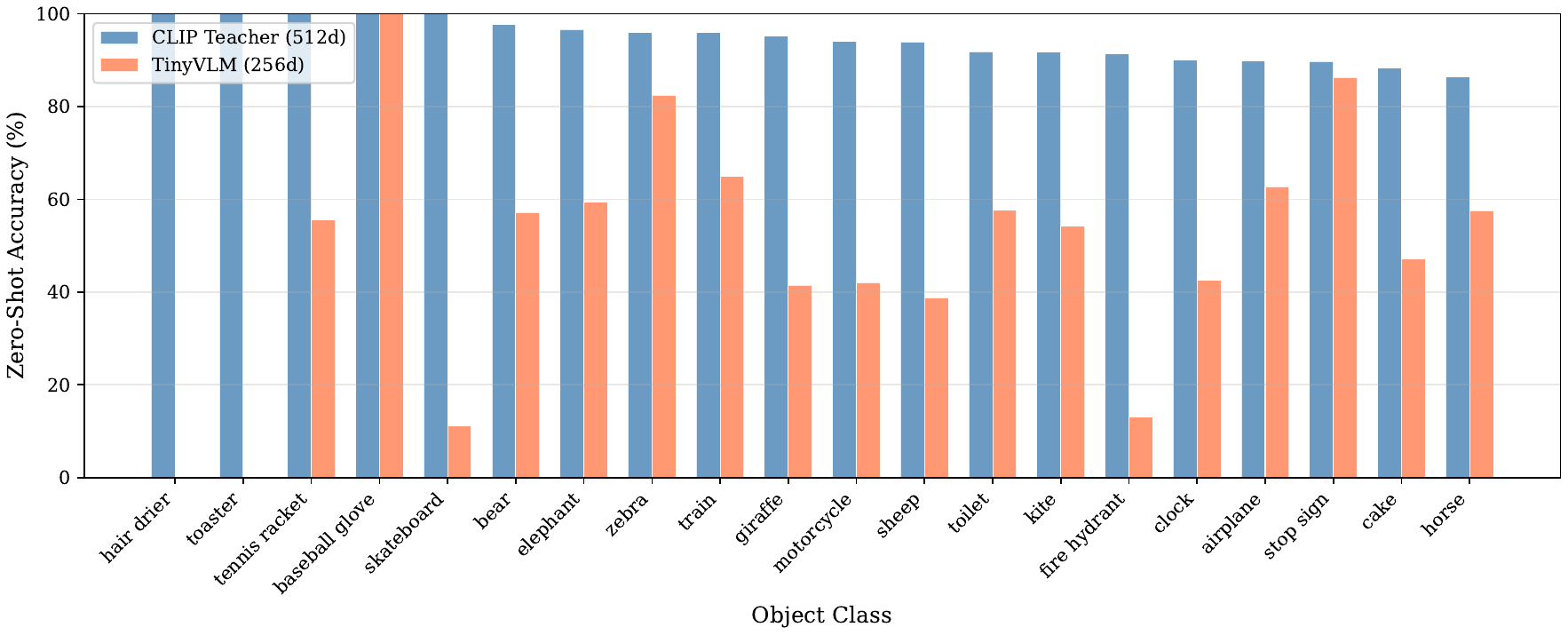}
\caption{Per-class zero-shot accuracy comparison between \ours{} (256-dim) and CLIP teacher (512-dim) on COCO val. Classes are sorted by teacher accuracy.}
\label{fig:perclass}
\end{figure}

\Cref{fig:perclass} shows the per-class zero-shot accuracy on COCO compared to the CLIP teacher.

\subsection{Embedding Visualization}
\label{app:embeddings}

\begin{figure}[tb]
\centering
\includegraphics[width=0.85\textwidth]{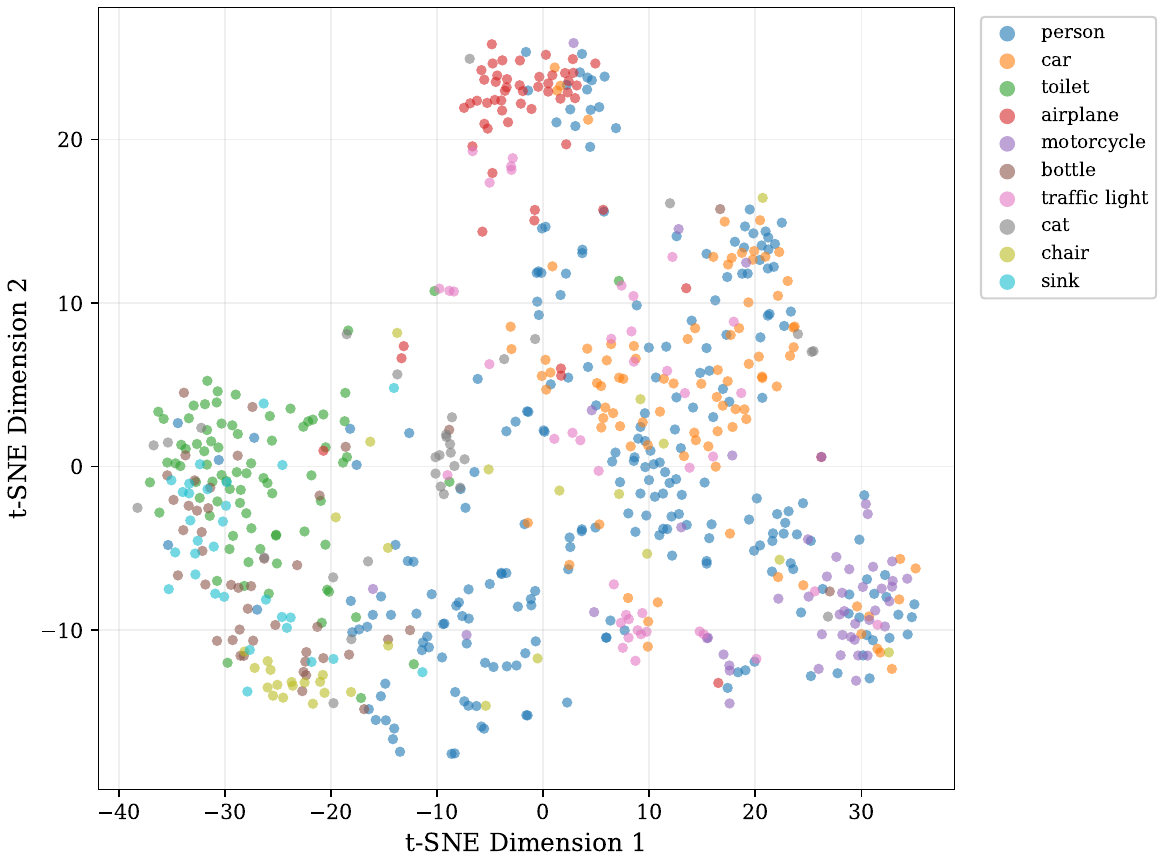}
\caption{t-SNE visualization of \ours{} image embeddings (64-dim) on COCO val, colored by class (top-10 most frequent). Embeddings show meaningful cluster structure despite aggressive dimensionality reduction.}
\label{fig:tsne}
\end{figure}

\Cref{fig:tsne} visualizes the learned embeddings using t-SNE.

\subsection{Dataset Comparison}
\label{app:dataset_comparison}

\begin{figure}[tb]
\centering
\includegraphics[width=0.9\textwidth]{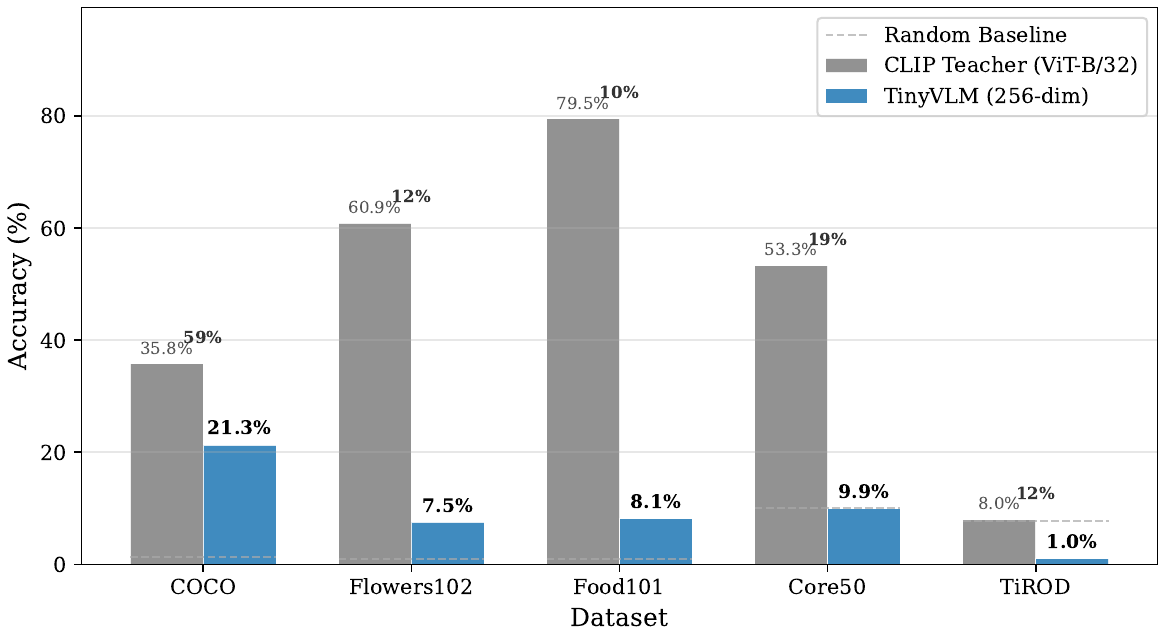}
\caption{Zero-shot accuracy (256-dim) across evaluation datasets compared to random baselines. Numbers above bars indicate improvement factor over random chance.}
\label{fig:dataset_comparison}
\end{figure}

\Cref{fig:dataset_comparison} compares \ours{} accuracy against random baselines across all evaluation datasets.

\subsection{Qualitative Predictions}
\label{app:qualitative}

\begin{figure}[tb]
\centering
\includegraphics[width=\textwidth]{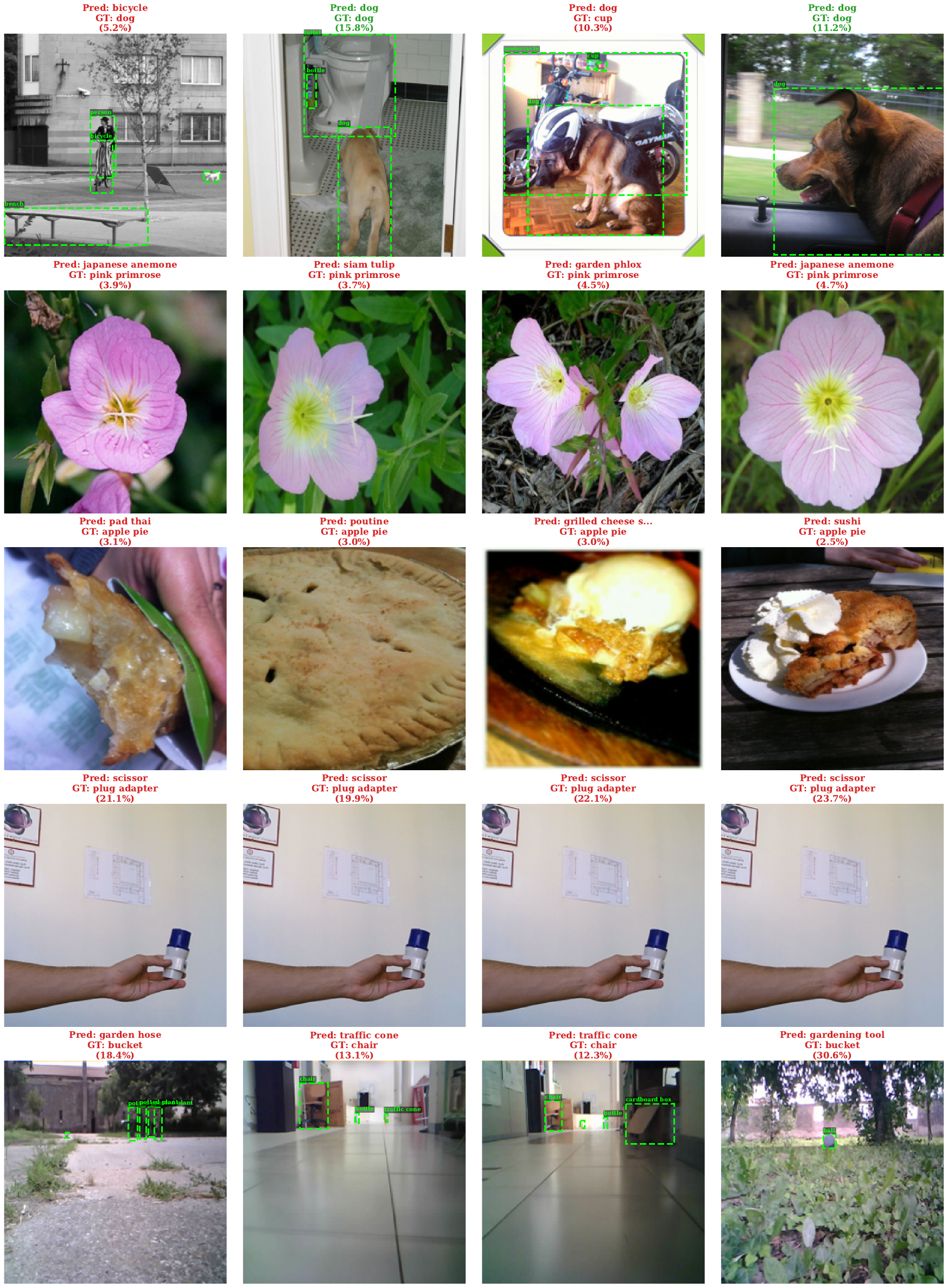}
\caption{Qualitative zero-shot predictions from \ours{} (256-dim) on sample images from each evaluation dataset. Correct predictions shown in green, incorrect in red, with confidence scores.}
\label{fig:qualitative}
\end{figure}

\Cref{fig:qualitative} shows example predictions from \ours{} across all five evaluation datasets.

\end{document}